\documentclass[twoside]{article}
\usepackage{ecj,palatino,epsfig,latexsym,natbib,amsmath}

\usepackage{booktabs}       
\usepackage{tabularx}
\usepackage{graphicx}
\usepackage{subcaption}
\usepackage{color}
\usepackage{qrcode}

\usepackage{hyperref}

\usepackage{xspace}
\newcommand{\eg}{e.g.,\xspace}
\newcommand{\ie}{i.e.,\xspace}

\usepackage{listings}
\usepackage{ulem}

\AtBeginDocument{\DeclareCaptionSubType{lstlisting}}
\newcommand{\lstin}[2][]{\lstinline[#1]|#2|}

\usepackage{framed}
\newenvironment{BNF}
  {\captionsetup{type=lstlisting}}
  {}
\usepackage[nounderscore]{syntax}
\usepackage[para,flushleft]{threeparttable}

\newcolumntype{L}[1]{>{\raggedright\let\newline\\\arraybackslash\hspace{0pt}}m{#1}}
\newcolumntype{C}[1]{>{\centering\let\newline\\\arraybackslash\hspace{0pt}}m{#1}}
\newcolumntype{R}[1]{>{\raggedleft\let\newline\\\arraybackslash\hspace{0pt}}m{#1}}

\usepackage{multirow}
\usepackage{dblfloatfix}
\usepackage{balance} 

\usepackage{color}

\begin{document}

\ecjHeader{x}{x}{xxx-xxx}{2024}{Evolving Assembly Code in an Adversarial Environment}{I. Maliukov, G. Weiss, O. Margalit and A. Elyasaf}
\title{\bf Evolving Assembly Code in an Adversarial Environment}  

\author{\name{\bf Irina Maliukov} \hfill \addr{irinamal@post.bgu.ac.il}\\ 
        \addr{Department of Computer Science, Ben-Gurion University of the Negev, Be'er Sheva, Israel}
\AND
       \name{\bf Gera Weiss} \hfill 
       \addr{geraw@bgu.ac.il}\\
        \addr{Department of Computer Science, Ben-Gurion University of the Negev, Be'er Sheva, Israel}
\AND
       \name{\bf Oded Margalit} \hfill \addr{odedm@post.bgu.ac.il}\\
        \addr{Department of Computer Science, Ben-Gurion University of the Negev, Be'er Sheva, Israel}
\AND
       \name{\bf Achiya Elyasaf} \hfill 
       \addr{achiya@bgu.ac.il}\\
        \addr{Department of Software and Information System Engineering, Ben-Gurion University of the Negev, Be'er Sheva, Israel}
}

\maketitle

\begin{abstract}
In this work, we evolve Assembly code for the CodeGuru competition. The goal is to create a survivor---an Assembly program that runs the longest in shared memory, by resisting attacks from adversary survivors and finding their weaknesses.
For evolving top-notch solvers, we specify a \textit{Backus Normal Form} (BNF) for the Assembly language and synthesize the code from scratch using \textit{Genetic Programming} (GP). We evaluate the survivors by running CodeGuru games against human-written winning survivors. 
Our evolved programs found weaknesses in the programs they were trained against and utilized them.
To push evolution further, we implemented memetic operators that utilize machine learning to explore the solution space effectively.
This work has important applications for cyber-security as we utilize evolution to detect weaknesses in survivors. The Assembly BNF is domain-independent; thus, by modifying the fitness function, it can detect code weaknesses and help fix them. 
Finally, the CodeGuru competition offers a novel platform for analyzing GP and code evolution in adversarial environments. To support further research in this direction, we provide a thorough qualitative analysis of the evolved survivors and the weaknesses found. 
\end{abstract}


\section{Introduction}
\begin{figure}
  \centering
\includegraphics[width=\linewidth]{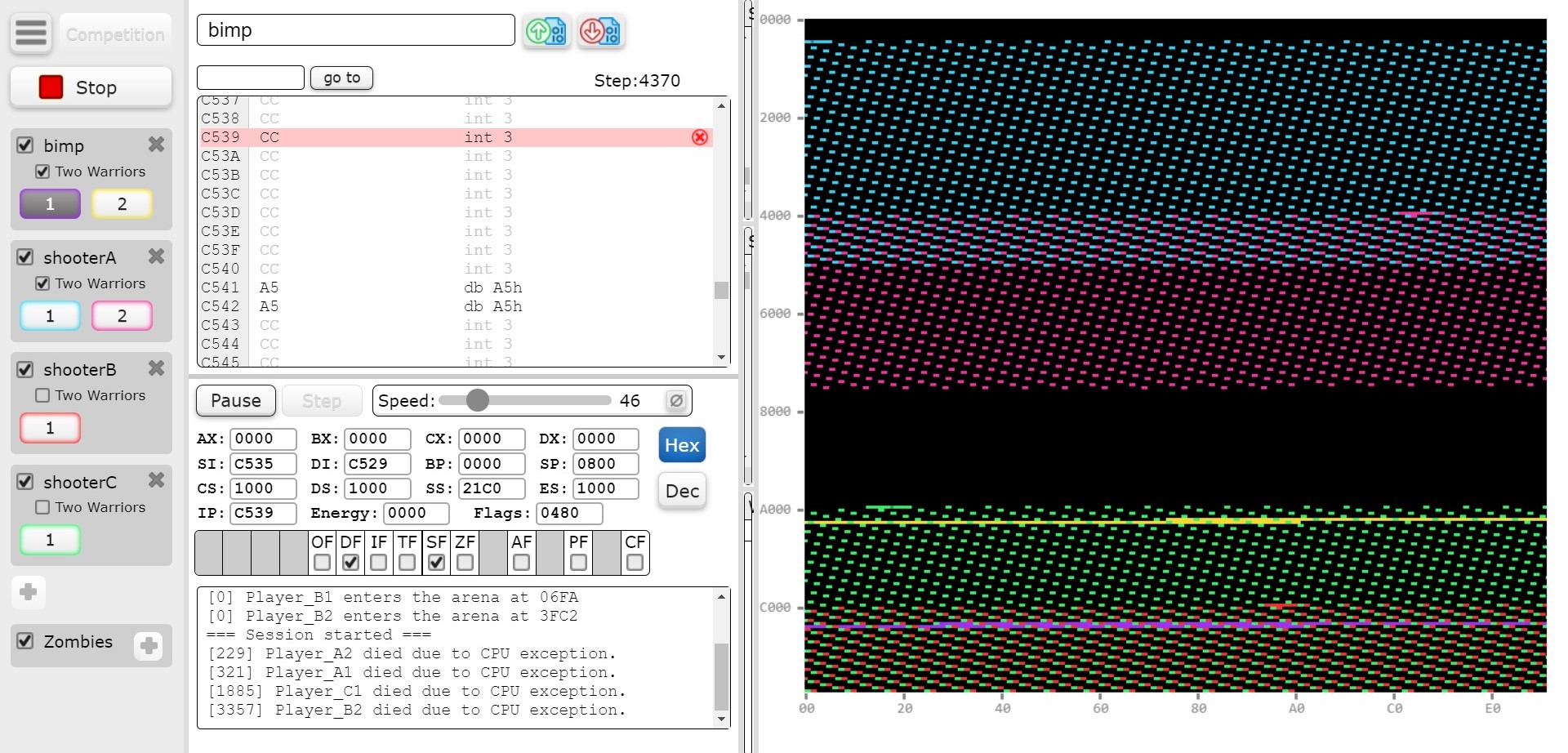} 
  \caption{The CodeGuru Xtreme game. On the left are the survivors; on the center is the code of the selected survivor; and on the right is the arena, \ie the memory status. Each survivor gets a different color in the arena, representing its written bytes.}
  \label{fig:CodeGuru}
\end{figure}

CodeGuru Xtreme~\cite{codeguru_repo} is a coding competition where short 8086 Assembly programs, called survivors, are loaded into a random address in a virtual computer memory arena. Their goal is to defeat all other survivors by staying the last program to run. An opponent is defeated when it runs an illegal command caused, \eg by overwriting its memory. A screen-shot of the game is depicted in \autoref{fig:CodeGuru}. Each survivor gets a different color in the arena, representing the bytes it wrote to the shared memory. We elaborate on the game in \autoref{sec:codeguru}.

In this work, we evolve winning survivors from scratch, \ie from randomly generated Assembly code, and without access to the source code of known survivors. For this task, we utilize \textit{Grammar-Guided Genetic Programming} (G3P)---an evolutionary computation technique that incorporates GP principles, employs context-free grammar and operates directly with tree-based representations. G3P allows us to evolve Assembly programs following grammar-type constraints. The goal of the individuals, embodied by the fitness function, is to overtake adversaries and win the game. The evolved code is represented using an \textit{Abstract Syntax Tree} (AST) based on matching Assembly \textit{Backus Normal Form} (BNF) we defined. A BNF is a meta-syntax notation for context-free grammar consisting of derivation rules. Since our BNF is general and domain-agnostic, the approach applies to generating Assembly programs for other domains and processors.

No previous work has been done on the CodeGuru Xtreme game, except for an undergraduate project~\cite{Darwin8086} and some early work on the ``Core War'' game~\cite{Andersen2001TheGE, corno2003exploiting}, which served as the basis CodeGuru Xtreme (see \autoref{sec:codeguru}). As elaborated in \autoref{sec:related-work}, some work has been done on the evolution of low-level languages (\ie Assembly and Java bytecode), and some work has been done to improve existing Assembly code. Note that improving existing code is a simpler task than generating new code from scratch, as the former's state space is much smaller~\cite{Petke2018GeneticImprovementSoftware, Banzhaf2018SomeRemarksCode}.

CodeGuru Xtreme competition has been running since 2005, with all past survivors publicly available. Thus, winning is considerably tricky and requires, among other qualities, a good understanding of the 8086 Assembly language.

This work combines a fitness approximation approach based on machine learning (ML) to overcome the expensive fitness evaluation. We utilize this approximation for developing memetic operators---genetic operators that incorporate local search to enhance the exploration, leading to significant improvements in the overall evolutionary computation.

Our work has implications for cyber-security. We utilize evolution to detect and exploit weaknesses in other survivors. Furthermore, understanding the Assembly language is a necessity for some viruses. By modifying the fitness function, our approach can be used for detecting weaknesses in code and help in fixing them, detecting suspicious adversarial code, or, on the contrary, can be intended to avoid security mechanisms by mutating the virus to avoid detection while keeping its functionality. In \autoref{sec:cyber}, we use the CodeGuru Xtreme game and GP for mimicking Assembly viruses and demonstrating how our method can be used for avoiding anti-virus detection.

The CodeGuru Xtreme competition provides a unique opportunity to analyze GP and code evolution in adversarial environments. In order to encourage further research in this area, we conducted a comprehensive qualitative analysis of the evolved survivors and identified their strengths and weaknesses. This analysis sheds light on the effectiveness of the evolved code and provides valuable insights for future improvements and advancements in the field.

The contributions of this work are:
\begin{itemize}
\item We develop a generic Assembly BNF.
\item We demonstrate, for the first time, how evolution can evolve Assembly code in an adversarial environment, completely from scratch.
\item We provide a qualitative analysis of the evolved survivors and the weaknesses found to facilitate further research in the field of cyber-security.
\item We demonstrate how CodeGuru and our approach can be utilized to mimic and explore virus behavior.
\end{itemize}

\section{Related Work}
\label{sec:related-work}
\paragraph{Low-level code evolution}
Several works have been done on low-level code evolution, some similar to Assembly, like Verilog (a hardware description language) and Java bytecode. 

~\cite{ulya2005automatic} used Grammatical Evolution (GE) for evolving a simple program of a one-bit full adder. In spite of the strong bias they employed, the success achieved was only about 5.7\%. 
~\cite{orlov2011flight} proposed a method for genetic improvement and repair of existing Java programs or any software that can be compiled into Java bytecode. Although Java bytecode resembles Assembly, it has a simplified representation and does not have direct memory access. This contrasts with Assembly, which has a very strong correspondence between its instructions, the architecture's machine code instructions, and memory. Orlov and Sipper seeded the initial population with copies of a single hand-crafted individual and improved it over generations while we evolve Assembly code from scratch.
\cite{Rosin2019Stepping} synthesized simple programs with loops from input/output examples. They targeted a simplified low-level language similar to Assembly, where each instruction consists of an opcode and a single operand. 

\paragraph{Limited Assembly code evolution}
Some works focused on constrained Assembly evolution---specific routines, predefined input and output tables, and manually written code parts. 
\cite{Serruto2017Automatic} applied multi-objective linear GP for the automatic generation of specific Assembly driver routines. The evolved programs did not contain jump instructions because they could form infinite loops, in contrast to our wish to include them in the generated code. The results showed that the automatically generated microcontroller code for specific tasks can compete with a human programmer with a smaller code size or faster execution. We aim to recreate this result within an adversarial environment.
Similarly, \cite{Ferrel2020Genetic} presented a methodology for writing Arduino programs using an automatic generator of Assembly language routines based on a cooperative co-evolutionary multi-objective linear GP. They decomposed the problem into sub-components, generating about 73\% of the program, and the remaining 27\%, which are the main program and initial configuration routines, are manually written.
In our case, we cannot break the goal of winning an opponent to sub-tasks without reading its code first, which we avoid. In addition, there is no clear way to represent a winning result using an input-output table. We need to evolve from scratch, a winning program, including jumps and loops, completely automatically.

\paragraph{Overview of Program Synthesis Methods}  
\cite{domonik2021recent} surveyed recent developments in program synthesis with evolutionary algorithms and found that the most influential approaches in the field are stack-based GP (usually PushGP), G3P, and Linear GP.
PushGP produces code only in the Push language and cannot be used outside of a Push interpreter, while G3P and Linear GP can produce source code in any language, including Assembly.

\cite{Pantridge2017On} compared the synthesis capabilities of PushGP and G3P over different programming languages. It was done on problems of two types: the first is usually approached using machine code or Assembly language (basic execution models problems), while the second is usually approached using high-level languages (general program synthesis benchmark suite). 
In the low-level field, the best-synthesized program was in TerpreT---a probabilistic programming language designed for inductive program synthesis. It was able to solve all the basic execution model problems, in contrast to PushGP, which succeeded in 6 out of 8. Sadly, there were no gathered results for G3P in this field. 
In the high-level field, G3P was able to solve almost as many problems as PushGP in the software synthesis benchmark suite.

All the related works present useful ideas for various research directions to achieve our goal of Assembly code evolution. Yet, none achieved total success in evolving independent Assembly programs from scratch. They all applied constraints and limitations on the programs, removed language features, or started from an initial given code. We aim to expand the above achievements.



\paragraph{GP usage in cyber-Security}
\cite{Oreilly2020Adversarial} stated that combining GP and competitive co-evolutionary algorithms enables evolving complex behaviors to be used as abstractions by adversaries. The paper presents the RIVALS framework, which serves as a testbed for computational modeling and simulation of the dynamics of networks under attack. \cite{Hu2020Optimal} describe the attack-defense relationship using LQRD (Logit Quantal Response Dynamics). They describe how, by analyzing the evolutionary equilibrium, we can obtain the optimal defense strategy and demonstrate it on WannaCry malware. 

In addition to the works that use GP to obtain strategies, others utilize it to evolve and modify malware. \cite{Noreen2009Evolvable} developed a feature representation of the Bagel malware family and, using GA and a few training samples, were able to evolve new unknown variants. 
\cite{Castro2019AIMED} used GP to evade malware detection by automatically finding code optimizations that, when injected into previously detected malware, result in misclassification of the malware scanner. They implemented their framework as a sandbox to track how the code samples behave. 
\cite{Murali2022Adapting} used novelty search in order to generate malware variants of greater diversity to evade detection. They used Assembly code samples and represented them in linear or graph representations. Generic Assembly code transformation functions are applied as operators: inserting fake instructions forced \texttt{jmp}, unreachable blocks, and conditional \texttt{jmp}s. They were able to evade over 98\% of popular scanners using this technique.

In our work, we demonstrate the ability to evade detection using GP with general Assembly grammar and domain-independent operators. Unlike the described above works, we do not create special operators for code obfuscation. We utilize CodeGuru as an adversarial game framework to evolve Assembly code, which is able to evade an adversary and is specially designed to overtake it based on a characterizing signature. This resembles diverse forms of viruses that avoid malware scanners.  

\section{CodeGuru Xtreme}
\label{sec:codeguru}
CodeGuru Xtreme by~\cite{codeguru_repo} is a coding competition based on ``Core War''---a 1984 programming game created by D. G. Jones and A. K. Dewdney~\cite{dewdney1984recreational}.
In CodeGuru Xtreme, short 8086 Assembly programs (at most 512 bytes long) of 16-bit commands, called survivors, are loaded into random space on a virtual computer memory arena of size 64KB. Each survivor is loaded to a random address with a stack of 2,048 bytes and a full set of registers. The distance between two survivors and the arena's edges is at least 1,024 bytes.
The last survivor alive is the winner and gets one point. If several survivors stay alive, the point is divided equally between all.
A survivor is disqualified if it runs an illegal command or attempts to access a memory address outside the arena or its stack. A survivor can be forced to run an illegal command due to a writing operation an opponent previously performed on its code or on an address it reads from. Therefore, performing various writings to memory enhances the chance of damaging opponents.
Each battle of the game runs for 200,000 rounds or until only one survivor is left, whichever comes first. In every round, the next command of each survivor is executed in a round-robin fashion. The order of the survivor's execution is changed randomly for each game. The memory arena image and scoreboard that monitors the game's progress are depicted in \autoref{fig:CodeGuru}. In most cases, each participant has two programs, called parts, that can collaborate together. Each part has its own registers and is loaded into a different place in memory, although they both share a stack. The parts are executed separately, one after the other. Their scores are joined together at the end of the battle and create the survivor's score. This allows the design of a survivor with two parts collaborating together via a shared stack or with two parts completely independently running; both designs are used as a force multiplier for maximizing the survivor's abilities. 
The game includes the following special commands: \texttt{WAITx4} increases the survivor's speed, allowing it to run several opcodes in a single round, \texttt{INT 0x86} writes 256 bytes into memory, and \texttt{INT 0x87} re-writes a pattern of 4 bytes. The special commands allow performing several rounds of actions in one round.

The CodeGuru competition has taken place every year since 2005 among outstanding high-school students. Each year, the level rises, with more sophisticated survivors written.
This work aims to evolve survivors that will win the top survivors of previous years by finding their weaknesses. Our goal is not to win the competition but rather to show that GP can be utilized for evolving code in an adversarial environment. Thus, we evolve a different survivor for each past survivor rather than evolving one survivor that takes them all.

\section{Method}
To evolve our survivors, we use Grammar-Guided Genetic Programming (G3P)---a technique that incorporates Genetic Programming principles, employs context-free grammar, often in a BNF form, and operates directly with tree-based representations. G3P allows evolving Assembly programs following grammar type constraints and a defined aim, overtaking adversary in this case. During the evolutionary process with G3P, the evolved code is represented using an AST based on matching Assembly BNF representation. 

We now elaborate on the different parts of the evolutionary process.

\subsection{Representation}
Each individual consists of \textit{two} programs, called \textit{parts} (see \autoref{sec:codeguru}), represented by an AST that follows a grammar defined by a BNF. Since Assembly is a symbolic programming language, it can be represented using it. The terminals are opcodes and operands, and the functions are structures in the language (see listings \ref{tab:1_terminal_set} and \ref{tab:t2_function_set} in the appendix). We define our types and derivation rules based on Assembly language constraints. For example, we define an unary command as a command consisting of an opcode that takes only one operand. Notably, except for a few CodeGuru special operators, our BNF is general and can match any 8086 Assembly code. There are Assembly commands that are not supported by the game's engine and were left out of the grammar to preserve legal programs.

\subsection{Fitness Function}
\label{sec:method:fitness}
We evaluate survivors' fitness by running a CodeGuru game of 200 battles with the selected human-written survivor the evolution performed against. The game's engine is an open-source Java program that outputs the final scores for each game. As previously explained, the score is one point given to the last survivor alive. If several survivors stay alive, the point is divided equally between all. We modified the engine to produce more information about each game, as elaborated by the fitness function that has four parts:

\textit{Engine score:} the survivor's average engine score in all played games. 
\begin{center}
$f_{\text{score}} = \frac{\sum_{i=1}^{\text{games}} \text{score}_i}{\text{games}}$
\end{center}

\textit{Lifetime:} the normalized average number of rounds the survivor stayed alive.
\begin{center}
$f_{\text{lifetime}} = 0.1 \log_{10} max\left(1, \frac{\sum_{i=1}^{\text{games}} \text{reached\_round}_i}{\text{games}}\right)$
\end{center}

\textit{Written bytes:} the normalized average number of new bytes the survivor wrote. That is, the writing was performed on a memory fragment, which was not written before, or that the last one to write in was not the survivor itself.
\begin{center}
$f_{\text{written\_bytes}} = 0.1 \log_{10} max\left(1, \frac{\sum_{i=1}^{\text{games}} \text{written\_bytes}_i}{\text{games}}\right)$
\end{center}

\textit{Writing rate:} the average writing rate of the survivor.
\begin{center}
$f_{\text{writing\_rate}} = 0.1 \frac{\sum_{i=1}^{\text{games}} \text{written\_bytes}_i}{max\left(1, \sum_{i=1}^{\text{games}} \text{reached\_round}_i\right)} \times \frac{1}{\text{games}}$
\end{center}

The first two parts encourage evolution to win the competitions and survive for longer periods (respectively). The last two parts encourage the evolution of programs that write in different memory places, which enhances the chance of damaging opponents. We refer to the score parameter as the most significant since it reflects the performance compared to the adversary. Nevertheless, the other parts are important for guiding the evolution towards the different sub-goals and discriminating the individuals. Division by 10 and $\log_{10}$ were used on the original values of lifetime and written\_bytes in order to normalize them to an easy-to-process range yet maintain the tendency they represent. We also defined a bloat weight parameter, which equals $10^{-5}$. It slightly lowers the fitness of large evolved trees in order to prevent them from bloating and yet allows large but powerful trees to evolve.

The fitness formula which performed the best was: 
\begin{equation*}
\begin{split}
  f = 2 f_{\text{score}} + 0.2 f_{\text{lifetime}} + 0.3 f_{\text{written\_bytes}} + 0.1 f_{\text{writing\_rate}} \\ - 10^{-5} max(\#part1\_nodes, \#part2\_nodes)  
\end{split}
\end{equation*}

It produces fitness values in the range of $[0, 2.5]$, which does not produce sharp deviations.
The chosen weights reflect the above-elaborated goals. The score is doubled due to its significance, resulting in a value in the range of $[0, 2]$. The additional parameters were multiplied by weights, resulting in a mutual sum of $0.5$ at most in order not to overshadow the score. According to conducted experiments, the writing\_rate parameter resulted in the smallest survivor's improvement, thus receiving the lowest weight. Experiments also showed that evolution was able to independently learn the importance of the survivor's lifetime, on the contrary to the importance of performed writings, which it learned seldom. Thus, high and medium weights were given to written\_bytes and lifetime parameters, respectively, to accelerate the improvement process.

\subsection{Genetic Operators}
\label{sec:method:operators}
We used Koza's standard mutation and crossover operators~\cite{koza1994genetic} that operate on the survivors' parts, which are represented as trees.
Specifically, we used the grow sub-tree (\ie sub-part) mutation and the exchange sub-tree (sub-part) crossover.
We added two more operators. The \textit{duplicate-tree} (part) mutation takes the best tree (part) of a survivor and replaces the second part with it. The \textit{exchange-trees} (parts) crossover replaces one of the trees (parts) of the first individual with one of the trees (parts) of the second.

\subsection{Improvements}
Below are several improvements to the basic setup that we tested.

\subsubsection{Random Generator Pattern}
We wish to add randomness to the BNF to allow our survivors to be unpredictable. Thus, we add Pseudo-Random Number Generator (PRNG) patterns to our BNF. Specifically, we added Linear Congruential Generator (LCG) and XOR-Shift Generators implementation to our grammar as shown in \autoref{table3_random_set} in the appendix.

\subsubsection{Fitness Approximation}
\label{sec:method:approx}
One of our main hold-backs is the fitness calculation time. Raising the Java engine, running 200 battles, and outputting the results into a file takes non-negligible time, which prolongs the evolution that itself requires many hours
to complete a run due to G3P's stochastic nature.
To handle this issue, we combined a machine learning model, as presented in~\cite{Tzruia2023Fitness}, that learns how to approximate the fitness value without evaluating the individuals. Since their approach is designed for vector-based representation, we first create a mapping between our AST representation into a float vector form. For that, we use the AST's size and the engine's parameters (elaborated in \autoref{sec:method:fitness}). 
For example, if an individual has reached a score of 0.3, a lifetime of 3.21827, wrote 0.69897 bytes in a rate of 0.01, and its larger tree contains 25 nodes, its vector representation will be [0.3, 3.21827, 0.69897, 0.01, 25].
During evolution, the individuals' vector representations and their actual fitness value are collected and used for training the ML model.
The ML model switches between actual and approximated fitness according to defined conditions. In the actual fitness phase, it performs learning on truly evaluated individuals until it reaches a sufficient level of correct prediction and switches to approximation. In our case, we chose the switch condition to be based on cross-validation (CV) error as it reaches 5\% of the maximal possible fitness, which equals 0.125 out of 2.5.

In each generation in the approximation phase, a certain percentage of the population is still evaluated regularly to maintain the model and save the accurate results of the best individual evolved so far. The percent which performed the best was 30\%. Less sampling resulted in non-convergence of the evolution, and more sampling increased the evolution time.

\subsubsection{Memetic Operators}
\label{sec:method:memtic}
As we will show, the use of fitness approximation dramatically reduced the computation time. Thus, we decided to utilize the fitness approximation to create smart memetic operators that use local search for further improving evolution.

Whenever we apply the basic genetic operators described above, we run them five times and choose the best individual according to the approximated fitness. 

\section{Experiments and Results}
\label{sec:results}
We carried out a comprehensive set of experiments aimed at winning the top human-written survivors. Our code is written in Python, using the EC-KitY toolkit~\cite{Sipper2023ECKity}. Our code and data are at~\href{https://github.com/irenamal/EC-KitY/tree/Assembly_code_generation}{Assembly code generation}. 
The code for the human-written survivors we compete with can be found at~\cite{codeguru_repo}.

Experiments were conducted on a shared cluster of 96 nodes and a total of 5,408 CPUs (the most powerful processors are AMD EPYC 7702P 64-core, although most have lesser specs). 64 CPUs and 150 GB RAM were allocated for each evolutionary run (against one human adversary) to parallelize the evaluation. In practice, each run without the fitness approximation took approximately two days. 

The specific hyperparameters utilized in the experiments and their chosen values are detailed in \autoref{tab:hyperparameters}. The population size was chosen to be 192 to optimize the need for diversity, considering the resources of 64 CPUs and assigned time per evolutionary run. The Grow mutation probability was chosen to be 0.7 as the experiments that were conducted showed a clear tendency to better results with a high mutation rate, yet this rate allows evolution to perform a significant learning process. 

\begin{table}
\caption{Evolutionary hyper-parameters.}
\label{tab:hyperparameters}
\centering
\begin{threeparttable}
\renewcommand{\arraystretch}{1.25}
\begin{tabular}[t]{R{200px}|p{150px}}
Representation & Grammar-based GP \\
Mutation & Grow sub-tree and duplicate tree \tnote{$\dagger$} \\
Recombination & Exchange of sub-trees and trees \tnote{$\dagger$} \\
Grow mutation probability & 0.7\\
Duplication mutation probability & 0.2\\
Exchange sub-tree recombination probability & 0.3\\
Replacement recombination probability & 0.2\\
Parent selection & Tournament with $k=4$\\
Survivor selection & Generational replacement\\
Population size & 192\\
Termination & 2,000 generations or convergence\newline with a winning strike
\end{tabular}
\renewcommand{\arraystretch}{1}
\begin{tablenotes}
\item[$\dagger$] The operators are described in \autoref{sec:method:operators}.
\end{tablenotes}
\end{threeparttable}
\end{table}

\begin{table}
\caption{Test average fitness and standard deviation over ten experiments of our best individuals against past years' winners. }
\label{table:test_results}
\centering
\begin{tabular}{c|c|c|c|c} 
\toprule
\textbf{Year} & \textbf{Human survivor} & \textbf{\#Wins} & \textbf{Avg. Engine's Score} & \textbf{SD} \\
\midrule
2006 & Zeus & 8/10 & 0.675 & 0.162 \\
2007 & HutsHuts & 10/10 & 0.960 & 0.048 \\
2008 & APOCALYPSE & 9/10 & 0.741 & 0.170 \\
2009 & XLII & 9/10 & 0.891 & 0.174 \\
2010 & FSM & 3/10 & 0.481 & 0.147 \\
2011 & Mamaliga & 9/10 & 0.738 & 0.132 \\
2012 & Zorg & 9/10 & 0.692 & 0.171 \\
2013 & Snake & 10/10 & 0.736 & 0.136 \\
2014 & IamAA & 6/10 & 0.478 & 0.220 \\
2014 & Paranoia & 9/10 & 0.890 & 0.190 \\
2015 & SilentError & 9/10 & 0.684 & 0.127 \\
2016 & LoudBugFix & 2/10 & 0.402 & 0.078 \\
2017 & Memz & 10/10 & 0.997 & 0.006 \\
2018 & Barvaz'sAngles & 10/10 & 0.991 & 0.008 \\
2019 & Nuki'sDemons & 5/10 & 0.666 & 0.286 \\
2020 & GreeniEs & 10/10 & 0.984 & 0.020 \\
2021 & BlocksOfGuru & 10/10 & 0.753 & 0.118 \\
2022 & TheHeapMen & 4/10 & 0.494 & 0.102 \\
\bottomrule
\end{tabular}
\end{table}

The operators were sequentially applied to individuals with different probabilities (\autoref{tab:hyperparameters}). 
The evolution was set to terminate when 2K generations were reached, or before, depending on whether convergence between best and average fitness values was achieved in addition to a monotonic non-increasing winning strike of 200 generations.

We repeated each experiment ten times to test consistency. Our individuals' average fitness and standard deviation against each of the past years' winners are in \autoref{table:test_results}. We consider an average engine score higher than 0.5 a winning result for our individual. Notably, evolution managed to evolve Assembly programs, which won almost 78\% of past years' human-written winners.

\begin{figure*}
\captionsetup{type=lstlisting}
\centering
\begin{sublstlisting}{0.45\textwidth}
\begin{lstlisting}
@start:
and cl, [bx + 0x68 + 0x104 + 0x246]
div WORD [bx]
l8293849:
rcl dl, cl
rcl ax, 1
rol si, cl
push WORD [si]
shl dh, cl
l8293850:
and WORD [di + 0x222], 0x92
wait
wait
!\textcolor{red}{\textbf{mov WORD} [di], 0x196}!
jns l8293850
@end:
\end{lstlisting}
\caption{Part 1}
\end{sublstlisting}
\hfill
\begin{sublstlisting}{0.45\textwidth}
\begin{lstlisting}
@start:
sub bh, [bx + 0x30 + @start]
div WORD [bx]
l8293858:
rol dl, cl
shr di, 1
!\sout{dw 0x144}! inc sp
!\sout{push ds}! add [0xDA80], bx
!\sout{sbb dl, 0x34}! xor al, 0xD3
!\sout{sar ax, cl}! clc
l8293859:
and WORD [si + 0x246 + 0x230], 0x264
push bx
!\textcolor{red}{\textbf{pop WORD} [di]}!
and WORD [si + 0x206 + 0x202 + 0x220 + 0x-14 + 0x-20 + 0x32], 0x144
jmp l8293859
@end:
\end{lstlisting}
\caption{Part 2}
\end{sublstlisting}
\caption{Evolved survivor against Zorg (2012 winner). The two parts of the evolved survivor utilized Zorg's Achilles' heel by writing data to a part of its program. Strike-through text denotes run-time changed code.}
\label{evolved_zorg}
\end{figure*}

\subsection{Qualitative Analysis of Evolved Survivors}
In this section, we inspect the code of the evolved solvers. The inspection reveals that the evolved survivors managed to win complex and long survivors using a relatively small code fragment. Although GP frequently evolves long code, sometimes only a small part of it is used in the program run flow and yet manages to win. As we will demonstrate, this shows how evolution found the Achilles' heel in the opponents' code and utilized it for its benefit.

\subsubsection{Utilizing Achilles' heel}
One of the clearest examples is Zorg---the 2012 winner. Zorg writes an important code fragment for its future run on memory address zero. The evolution process noticed it in about 100 generations and overridden this memory by addressing \lstin{di}, which holds the value zero, depriving Zorg of winning (see line 14 in both parts of \autoref{evolved_zorg}, which includes the \lstin{dw} translation to Assembly commands and the effect on the following commands). Zorg's code is significantly longer and more complicated than the evolved fragment that overtook it. The evolved survivor manages to win Zorg in about 70\% of the battles in a game (according to the average engine's score) despite the weakness finding due to the randomness in the game's execution order.

\subsubsection{Concentrated vs. Scattered Memory Writes}
\label{scattered}
\begin{figure}
    \centering
    \includegraphics[width=0.8\linewidth]{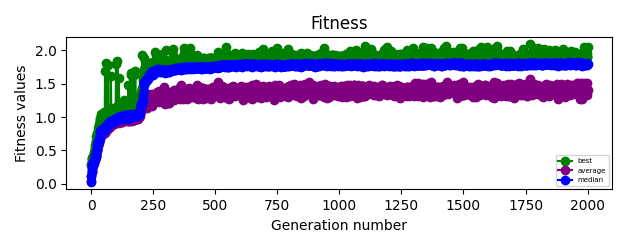}
    \caption{BlocksOfGuru (2021)}
    \label{fig:hist1}
\end{figure}

\begin{figure}
  \centering
    \includegraphics[width=\linewidth]{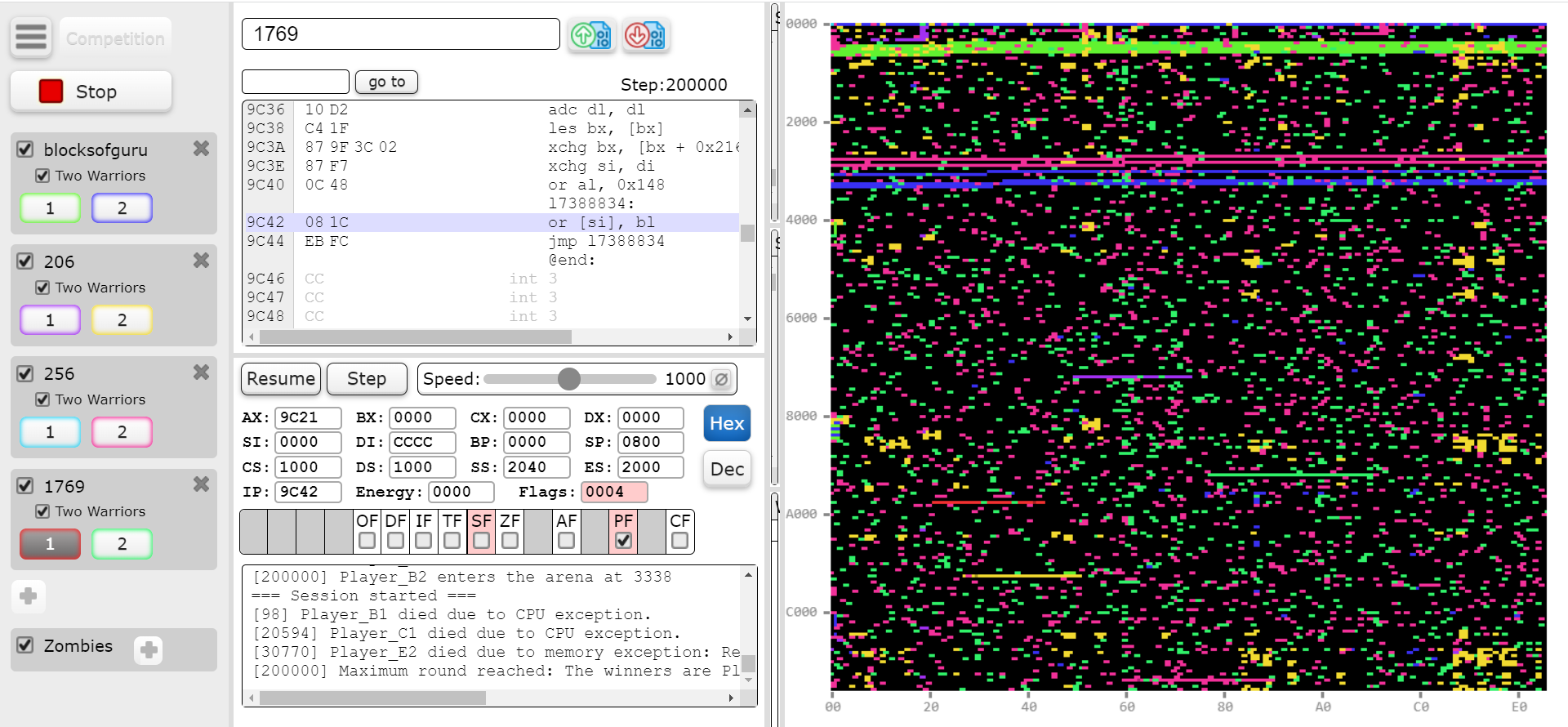}
  \caption{The memory image of BlocksOfGuru vs. best individuals from generations 206, 256, and 1,769. The scattered green and pink dots are memory bytes written by the 1,769 and 256 individuals, respectively. The concentrated yellow dots are memory bytes written by the 206 individual.}
  \label{fig:joinedgame}
\end{figure}

During evolution, we noticed several spikes in the best fitness. For example, when training against BlocksOfGuru, there were spikes in the fitness of the best individuals in generations 206 and 256 (see \autoref{fig:hist1}). To analyze these spikes, we ran a game with BlocksOfGuru against these individuals and the overall best individual (from generation 1,769). The results and memory image are depicted in \autoref{fig:joinedgame}.
We can see that most memory writes were made by the second part of the 1,769 and 256 individuals that cover the arena with scattered green and pink dots. 206's second part performed less, yet a significant number of writes in yellow are concentrated in several areas. All of the first part performed little to no new memory writes. Inspecting their cleared runnable code (Listings \ref{lst:BlocksOfGuru:a} and \ref{lst:BlocksOfGuru:b}) reveals that the first parts of 256 and 1,769 run in the loop written in their bottom, which keeps the individual alive but does not perform attacking actions---as seen in the lack of their color in the arena. In 206, both parts run in a loop due to \lstin{jmp ax} at the end, which jumps into the beginning of the code. Most memory writes of all individuals are performed using addressing \lstin{si}, yet with adding different constants to \lstin{si}. 256 and 1,769 use special constants like 65,535 and @start, while 206 does not. The first two exceed the bounds of word data, and the computation is thus written to an address defined as the special constant modulo $2^{16}$, which results in scattered writes. The evolutionary process discovered that scattered writing has more chances to encounter adversary code, as reflected in their higher scores, and it is reflected in runs against other survivors as well. 

\begin{figure*}
\captionsetup{type=lstlisting}
\centering
\begin{sublstlisting}[b]{0.23\textwidth}
\begin{lstlisting}
@start:
not WORD [bx+65535]
cmc
and [si+0x252],di
xchg bx,[si+0x051E]
jnc l268089
l268089:
add [si+0x0802+65535],bx
sbb cl,0x-14
loop l268090
l268090:
mov [si+@start+0x03B4],cx
rcr si,1
and [si],ax
jmp ax
@end:
\end{lstlisting}
\caption{206 part 1}
\end{sublstlisting}
\hfill
\begin{sublstlisting}[b]{0.22\textwidth}
\begin{lstlisting}
@start:
inc WORD [bx+65535]
cmc
and [si+0x252],cx
xchg bx,[si+0x051E]
jnc l268094
l268094:
add [si+0x0814+65535],bx
sbb cl,0x124
loop l268095
l268095:
mov [si+0x152],cx
rcr si,1
and [si],ax
jmp ax
@end:
\end{lstlisting}
\caption{206 part 2}
\end{sublstlisting}
\hfill
\begin{sublstlisting}[b]{0.19\textwidth}
\begin{lstlisting}
@start:
dec WORD [bx+65535]
cmc
and [si+0x252],dx
xchg bx,[di+0x05FC]
l376145:
sub [si],cl
jmp l376145
@end:
\end{lstlisting}
\caption{256 part 1}
\end{sublstlisting}
\hfill
\begin{sublstlisting}[b]{0.24\textwidth}
\begin{lstlisting}
@start:
inc WORD [bx+0x260+65535]
cmc
and [si+0x252],di
xchg bx,[si+0x0802+65535]
jnc l376150
l376150:
add [si+0x0802+65535],bx
sbb cl,0x-14
loop l376151
l376151:
mov [si+@start+0x03B4],cx
rcr si,1
and [si],ax
jmp ax
@end:
\end{lstlisting}
\caption{256 part 2}
\end{sublstlisting}
\caption{Comparing the code of best individuals against the BlocksOfGuru survivor.}
\label{lst:BlocksOfGuru:a}
\end{figure*}

\begin{figure}
\captionsetup{type=lstlisting}
\centering
\begin{sublstlisting}[b]{0.38\linewidth}
\begin{lstlisting}
@start:
xchg bp,[bp+0x0130]
and [di+0x072C],dx
xchg di,[di+0x05FC]
l7388834:
or [si],bl
jmp l7388834
@end:
\end{lstlisting}
\caption{1,769 part 1}
\end{sublstlisting}
\qquad
\begin{sublstlisting}[b]{0.5\linewidth}
\begin{lstlisting}
@start:
inc WORD [bx+0x260+65535]
cli
and [si+0x252],di
xchg bx,[si+0x0802+65535]
jnc l7388849
l7388849:
add [si+0x0802+65535],bx
sbb cl,0x-14
loop l7388850
l7388850:
mov [si+@start+0x03B4],cx
rcr si,1
and [si],ax
jmp ax
@end:
\end{lstlisting}
\caption{1,769 part 2}
\end{sublstlisting}
\caption{Comparing the code of best individuals against the BlocksOfGuru survivor.}
\label{lst:BlocksOfGuru:b}
\end{figure}

\begin{figure*}
  \centering
  \begin{subfigure}{0.45\textwidth}
    \includegraphics[width=\linewidth]{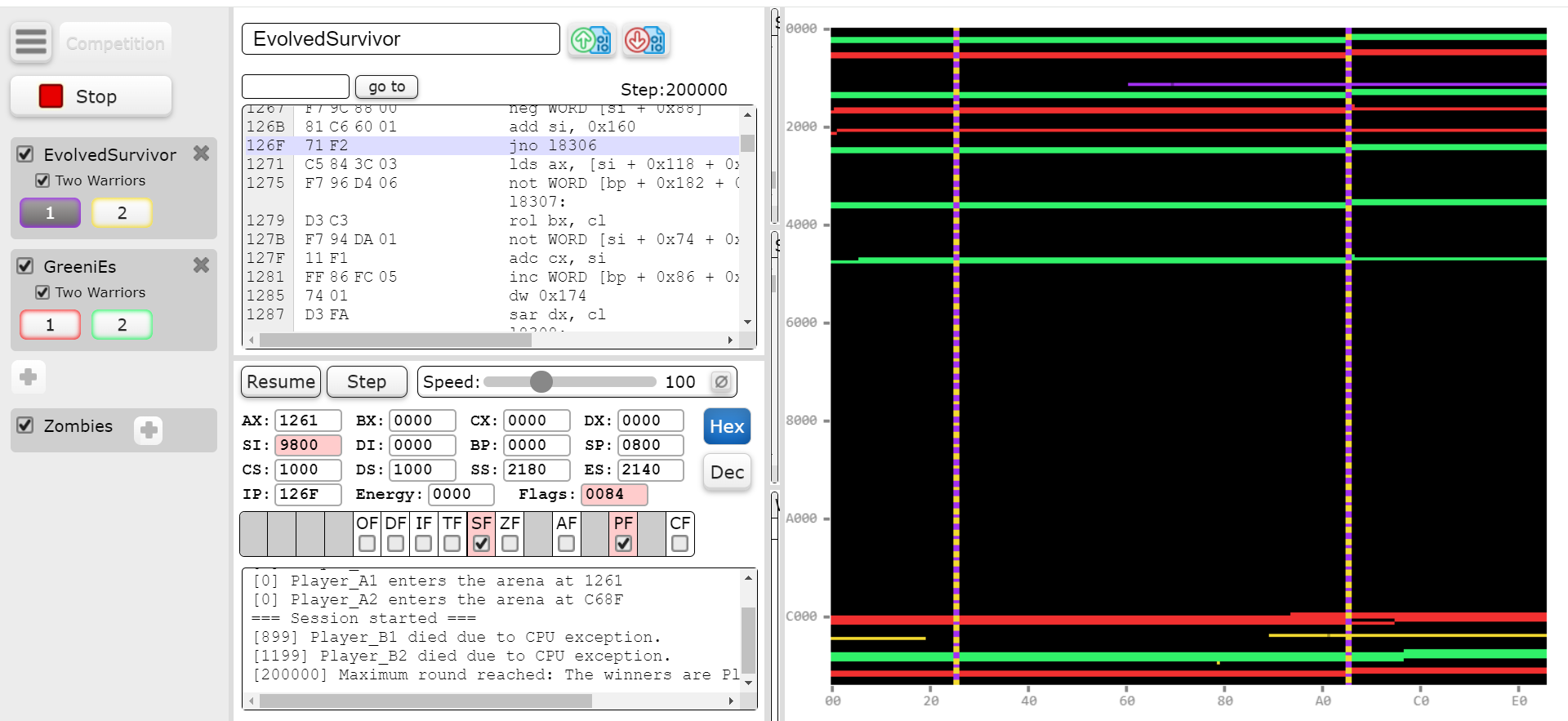}
    \caption{Evolution vs. GreeniEs (2020)}
    \label{fig:mem_after2}
  \end{subfigure}
  \hfill
  \begin{subfigure}{0.45\textwidth}
    \includegraphics[width=\linewidth]{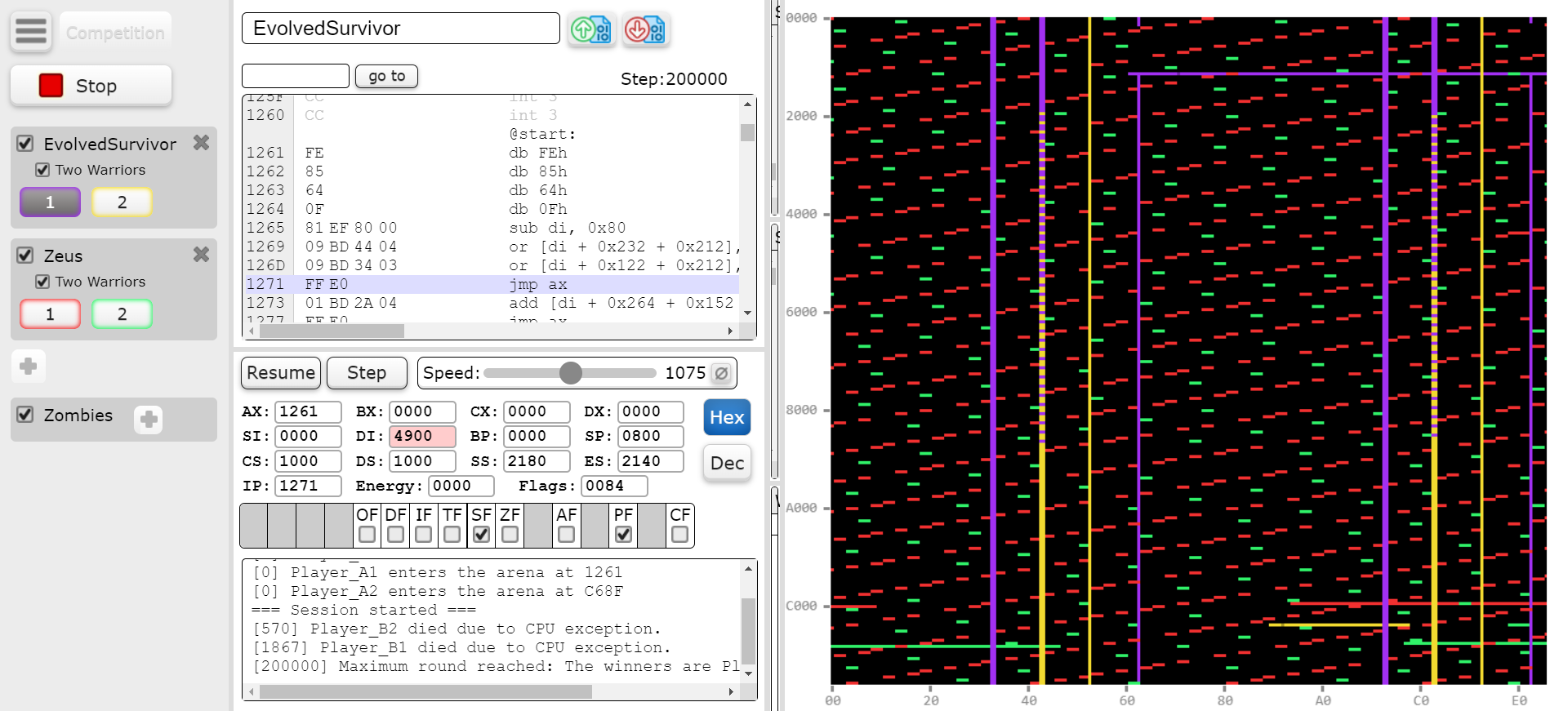}
    \caption{Evolution vs. Zeus (2006)}
    \label{fig:mem_after1}
  \end{subfigure}
  \caption{Vertical vs. horizontal memory write. Horizontal writings of evolved survivors (in purple and yellow) ``cut'' the human-written adversaries that write in horizontal lines (green and red) by writing on their code before they reach their code.}
  \label{fig:vertical}
\end{figure*}

\subsubsection{Vertical vs. Horizontal Memory Writes}
\label{vertical}
Another interesting pattern we detected in evolved survivors was writing vertical bytes into memory. That is, sequentially writing a byte every 256 bytes, creating a vertical line in the memory state. This contrasts with human-written survivors, who usually write memory in horizontal lines (consecutive bytes). As a result, the evolved survivors were able to ``cut'' the adversary by writing on its code before the adversary reached its code. This is depicted, for example, in the memory state of our evolved individuals against Zeus (2006) and GreeniEs (2020) (see \autoref{fig:vertical}). The evolved survivors, in purple and yellow, write their code in vertical lines, thus cutting their adversaries' horizontal writing (in red and green). Writing vertically assures reaching the opponent's code faster because most submitted survivors take advantage of the maximum allowed code size, thus filling at least one memory row entirely. Therefore, filling one or a few columns will be faster than filling complete rows. A similar pattern was found in many runs against other survivors.

The presented patterns of utilizing weak points, scattered and vertical writings are expressed in a significant part of the performed evolutionary runs against all the survivors, although not all the runs of each survivor have used the same pattern.

\subsection{Random Generator Pattern}
The original program had difficulties overtaking a few previous years' winners, specifically FSM 2010, IamAA 2014, LoudBugFix 2016, and TheHeapMen 2022, resulting in an average score lower than 0.5. We assumed that the BNF extension of random generator patterns may improve our evolved survivors. We ran the evolution against the above adversaries again for ten runs each.

As \autoref{tab:test_random_results} shows, using randomness improved the number of games evolution won and the average score in three out of four cases. The majority of the best-evolved individuals contained at least one of the random patterns. However, in some, the pattern appears in an unreachable code segment or outside the loop, meaning it only executes once. We believe it helped the evolution process, even though the winning survivor does not actively use it.

\begin{figure}
  \centering
    \includegraphics[width=\linewidth]{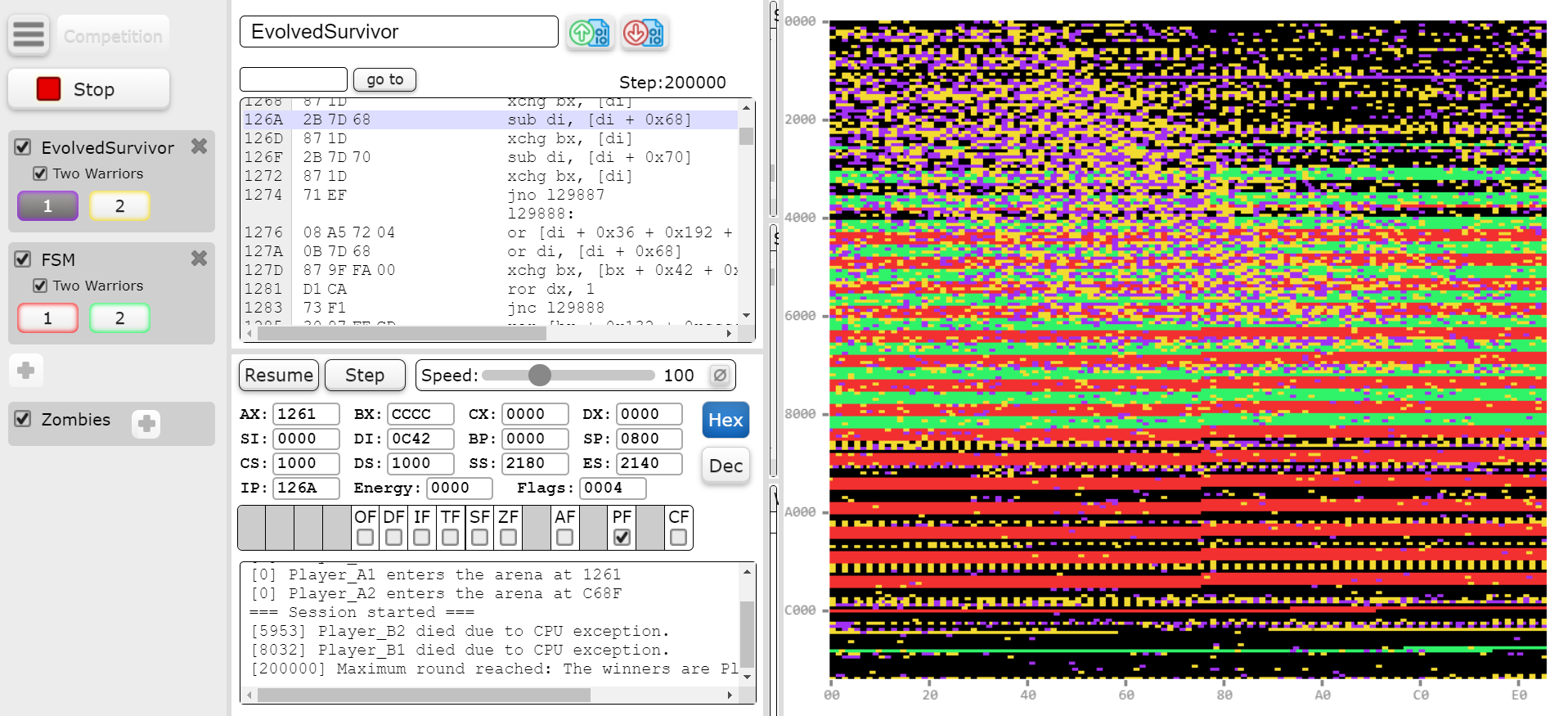}
    \caption{Scattered, horizontal, and vertical writing in the survivor, which evolved using randomness patterns against FSM (2010).}
    \label{fig:fsm_rand}
\end{figure}

The use of randomness enhanced the use of scattered writing patterns for some of the survivors and evolved a combined horizontal-vertical writing pattern for others, in contrast to the vertical-only pattern. The random pattern allowed for the combination of the described patterns together, resulting in scattered writing in horizontal lines that expand vertically, as seen in \autoref{fig:fsm_rand}. We can see the yellow and purple memory cells that are being filled horizontally at the beginning. Afterward, the created lines expand vertically, and everything is done using scattered writing. 

\begin{table}
\centering
\begin{tabular}{c|c|c|c|c|c} 
\toprule
\multirow{2}{*}{\textbf{Year}} & \multirow{2}{*}{\textbf{Human Survivor}} & \multicolumn{2}{c|}{\textbf{w/o Random}} & \multicolumn{2}{c}{\textbf{w/ Random}} \\\cline{3-6}
&&\textbf{Avg. Score} & \textbf{\#Victories} & \textbf{Avg. Score} & \textbf{\#Victories} \\ 
\midrule
2010 & FSM & 0.481 & 3/10 & 0.483 & 5/10 \\
2014 & IamAA & 0.478 & 6/10 & 0.377 & 3/10 \\
2016 & LoudBugFix & 0.402 & 2/10 & 0.483 & 6/10 \\
2022 & TheHeapMen & 0.494 & 4/10 & 0.496 & 5/10 \\
\bottomrule
\end{tabular}
\caption{Test game results with and without randomness.}
\label{tab:test_random_results}
\end{table}

\subsection{Fitness Approximation and Memetic Operators}
For the ML-based fitness approximation approach (\autoref{sec:method:approx}), we used the Ridge model over Lasso (see \cite{Tzruia2023Fitness}) as it helps reduce overfitting that results from model complexity and doesn't set the value of the coefficient to absolute zero. The complete fitness-approximation hyperparameters are given in \autoref{tab:hyperparameters_ML}, and they are used with the parameters described in \autoref{tab:hyperparameters}.

The combination of the approximation model in our evolution led to about a 30\% decrease in total evolution time while preserving the winning results achieved before (see \autoref{table:test_results}). For example, the evolution vs Mamaliga (2011) previously took about 77 hours and was decreased to 56 hours with the same amount of generation until winning was achieved.

\begin{table}
\caption{ML hyper-parameters.}
\label{tab:hyperparameters_ML}
\centering
\begin{threeparttable}
\renewcommand{\arraystretch}{1.25}
\begin{tabular}[t]{R{95px}|p{125px}}
switch\_condition & CV Error \\
switch\_threshold & 0.125 \\
sample\_rate & 30\% \\
model & Ridge \\
alpha & 0.3\\
gen\_weight & sqrt\\
\end{tabular}
\renewcommand{\arraystretch}{1}
\begin{tablenotes}
\end{tablenotes}
\end{threeparttable}
\end{table}

As the approximated fitness was relatively accurate, we decided to utilize it against the human-written survivors we were not able to overtake \autoref{tab:test_random_results}. We did that using the memetic operators described in \autoref{sec:method:memtic}, running 10 evolutionary processes against each opponent to establish our results, similar to the method described in \autoref{sec:method:fitness}. We also used the random-pattern extension in these experiments.

As \autoref{tab:test_approx_results} shows, there was a great improvement, and a win was achieved against all human-written survivors we were not able to win before. Furthermore, in all games in the test phase that resulted in winning, they achieved the best possible score of 1.0. They did not win all games, hence the average score of 0.8 and 0.9.

\begin{table}
\caption{Test game results with the memetic operators and random patterns.}
\label{tab:test_approx_results}
\centering
\begin{tabular}{c|c|c|c|c|c} 
\toprule
\multirow{2}{*}{\textbf{Year}} & \multirow{2}{*}{\textbf{Human Survivor}} & \multicolumn{2}{c|}{\textbf{w/o memetic operators}} & \multicolumn{2}{c}{\textbf{w/ memetic operators}} \\\cline{3-6}
&&\textbf{Avg. Score} & \textbf{\#Victories} & \textbf{Avg. Score} & \textbf{\#Victories} \\ 
\midrule
2010 & FSM & 0.481 & 3/10 & 0.8 & 8/10 \\
2014 & IamAA & 0.478 & 6/10 & 0.9 & 9/10 \\
2016 & LoudBugFix & 0.402 & 2/10 & 0.9 & 9/10 \\
2022 & TheHeapMen & 0.494 & 4/10 & 1 & 10/10 \\
\bottomrule
\end{tabular}
\end{table}

During the experiment, the ML model was able to learn the parameter weights in each run. The highest weight was given consistently to the score parameter, as expected, and the size parameter was approximately ignored in all cases. We also noticed that the writing\_rate weight varied among the survivors, while the other weights had close values.


\section{Application to Cyber Security}
\label{sec:cyber}
Signatures are widely used in anti-virus programs as a technique to detect known malware. They work by comparing specific patterns or characteristics of files against a database of known malware signatures. If a match is found, the anti-virus program can take appropriate action to quarantine or remove the malicious file. To bypass those techniques, malware developers have devised methods to alter their code in a manner that avoids detection by the signatures while still maintaining the malware's functionality. This perpetuates the ongoing battle between protectors and attackers in an endless cycle.

The adversarial environment of CodeGuru is an interesting platform for analyzing the ability of evolution to overcome tools designed especially against it, like the ability of malware developers to overcome signatures.
As mentioned in \autoref{sec:codeguru}, there is a special opcode \texttt{INT 0x87} in the game. It looks for a 4-byte sequence identical to the values stored in the registers \texttt{AX:DX} and replaces them with the values stored in \texttt{BX:CX}. The search is performed in the memory starting from address \texttt{DI:ES} moving up or down, determined by the direction flag. Therefore, if a command from a survivor is stored in \texttt{AX:DX}, it can be found and replaced by the adversary to an illegal command stored in BX:CX, causing the survivor to run it and be disqualified. This simulates the identification of malware by a signature.

We conducted an experiment utilizing the \texttt{INT 0x87} command to simulate signature-based anti-virus. The adversaries chosen were XLII (2009) and GreeniEs (2020), both of which we had previously defeated and that incorporate \texttt{INT 0x87} in their code. We subjected them to a complete evolutionary process, with a minor adjustment of halting evolution once the evolved population achieved a consistent win, indicated by an average fitness of 1.1, signifying a winning score (refer to \autoref{sec:method:fitness}). During this halt, we manually altered the adversaries' \texttt{AX:DX} values to match the commands relied upon by the best-evolved individuals. Subsequently, we resumed the evolution, introducing the evolved population to a bespoke adversary.

\begin{figure*}
  \centering
  \begin{subfigure}{0.47\textwidth}
    \includegraphics[width=\linewidth]{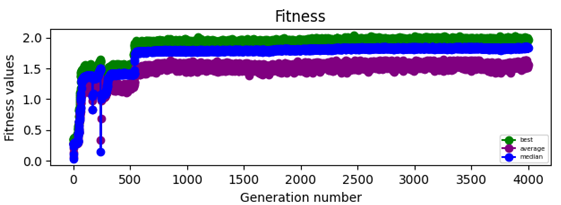}
    \caption{Evolution vs. modified XLII (2009)}
    \label{fig:antivirus1}
  \end{subfigure}
  \hfill
  \begin{subfigure}{0.47\textwidth}
    \includegraphics[width=\linewidth]{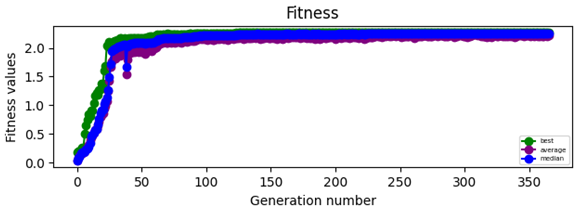}
    \caption{Evolution vs. modified GreeniEs (2020)}
    \label{fig:antivirus2}
  \end{subfigure}
  \caption{The fitness dropped in generation 241 and 38, respectively, after the tailor-made adversary was created.}
  \label{fig:antivirus}
\end{figure*}

In both cases shown in \autoref{fig:antivirus}, the fitness initially dropped in the following generation due to encountering an improved adversary. However, as evolution progressed, it managed to recover and even improve the fitness results to levels achieved before. The evolutionary process successfully replaced the targeted commands with different ones that exhibited similar behavior. These results demonstrate the capability of evolution to confront tools specifically designed to counter it. Furthermore, the use of a domain-independent Assembly grammar, coupled with the prevalence of malware written in Assembly, suggests that these findings could be leveraged to modify malware for evading signature-based anti-virus systems.

\section{Conclusion}
This work focused on evolving Assembly code for the CodeGuru competition with the objective of creating a survivor program that can run the longest in shared memory while withstanding attacks from adversary survivors. By defining a Backus Normal Form (BNF) for the Assembly language and employing Genetic Programming (GP) to synthesize the code, the study aimed to develop top-notch solvers. The evaluation of the evolved programs involved running CodeGuru games against human-written winning survivors, leading to the identification and exploitation of weaknesses in the opponent programs.

To enhance the evolutionary process, we presented three enhancements. Specifically, we added random patterns to the BNF to allow our survivors to be less predictable; we dramatically cut the overall training time approximating the fitness using a machine-learning (ML) approach; and finally, by developing ML-based memetic operators that allowed our survivors to outperform all pervious-years human-written survivors.

The significance of this research extends to the realm of cyber-security, where the evolved Assembly programs were adept at detecting vulnerabilities in the opponent survivors, showcasing the potential for utilizing evolutionary algorithms to identify and rectify code weaknesses. Our domain-independence Assembly BNF opens up possibilities for adapting the approach to various contexts by adjusting the fitness function to target specific code vulnerabilities.

Moreover, the CodeGuru competition serves as a valuable platform for studying Genetic Programming and code evolution within adversarial environments. The thorough qualitative analysis conducted on the evolved survivors and the vulnerabilities uncovered contributes to the body of knowledge in this area. By shedding light on the efficacy of evolutionary techniques in enhancing program robustness and resilience against attacks, this research paves the way for further investigations into evolutionary strategies for cybersecurity applications and code optimization in competitive settings.

\newpage

\clearpage
\appendix

\section{Our Grammar}
\begin{BNF}
\begin{framed}
\begin{grammar}
<reg> ::= `ax', `bx', `cx', `dx', `si', `di', `bp', `sp' 

<half\_reg> ::= `ah', `al', `bh', `bl', `ch', `cl', `dh', `dl'

<addres> ::= `[bx]', `[si]', `[di]', `[bp]'

<pop\_reg> ::= `ax', `bx', `cx', `dx', `si', `di', `bp', `WORD [bx]', `WORD [si]', `WORD [di]', `WORD [bp]', `ds', `es'

<push\_reg> ::= `ax', `bx', `cx', `dx', `si', `di', `bp', `WORD [bx]', `WORD [si]', ` WORD [di]', ` WORD [bp]', `ds', `es', `cs', `ss'

<const> ::= [(2*i) for i in range(-10, 133)]', `@start', `@end', `65535', `0xcccc'

<op> ::= `nop', `stosw', `lodsw', `movsw', `cmpsw', `scasw', `pushf', `popf', `lahf', `stosb', `lodsb', `movsb', `cmpsb', `scasb', `xlat', `xlatb', `cwd', `cbw', `cmc', `clc', `stc', `cli', `sti', `cld', `std'

<op\_single> ::= `div', `mul', `inc', `dec', `not', `neg'

<op\_double> ::= `cmp', `mov', `add', `sub', `and', `or', `xor', `adc', `sbb', `test'

<op\_jmp> ::= `jmp', `jcxz', `je', `jne', `jp', `jnp', `jo', `jno', `jc', `jnc', `ja', `jna', `js', `jns', `jl', `jnl', `jle', `jnle', `loopnz', `loopne', `loopz', `loope', `loop'

<op\_rep> ::= `rep', `repe', `repz', `repne', `repnz'

<op\_function> ::= `call', `call near', `call far'

<op\_special> ::= `wait wait wait wait', `wait wait', `int 0x86', `int 0x87'

<op\_pointer> ::= `lea', `les', `lds'

<op\_ret> ::= `ret', `retn', `retf', `iret'

<op\_push> ::= `push'

<op\_pop> ::= `pop'

<op\_double\_no\_const> ::= `xchg'

<op\_shift> ::= `sal', `sar', `shl', `shr', `rol', `ror', `rcl', `rcr'

<section> ::= ` ' 
\end{grammar}
\end{framed}\vspace{-1em}
\caption{Terminals definitions}\vspace{2em}
\label{tab:1_terminal_set}
\end{BNF}

\begin{BNF}
\begin{framed}
\begin{grammar}
<section> ::= <label> <section> <backwards\_jump> <section>
\alt <label> <section> <backwards\_jump>
\alt <section> <forward\_jmp><section> <label> <section>
\alt <label> <section> <call\_func> <backwards\_jump> <label> <section> <return>
\alt <op\_double> <reg>  <reg | const | address> <section>
\alt <op\_double> <address> <reg | half\_reg> <section>
\alt <op\_double> <half\_reg> <half\_reg | const | address> <section>
\alt <op\_double> <WORD | BYTE> <address> <const> <section>
\alt <op\_pointer> <reg> <address> <section>
\alt <op\_double\_no\_const> <reg> <reg | address> <section>
\alt <op\_double\_no\_const> <half\_reg> <half\_reg | address> <section>
\alt <op\_single> <reg | half\_reg> <section>
\alt <op\_single> <WORD | BYTE> <address> <section>
\alt <op\_function> <address> <section>
\alt <op | op\_special> <section>
\alt <op\_rep> <op> <section>
\alt <op\_push> <push\_reg> <section>
\alt <op\_pop> <pop\_reg> <section>
\alt `jmp' <reg | address> <section>
\alt `dw 0x'<const> <section>
\alt <op\_shift> <reg | half\_reg> <cl | 1> <section>

<call\_func> ::= `call l'<const> <section>

<return> ::= <op\_ret> <section>

<label> ::= `l'<const> <section>

<forward\_jump> ::= <op\_jmp> `l'<const> <section>

<backwards\_jump> ::= <op\_jmp> `l'<const>`-1' <section>

<address> ::= [<address> + <const>] 
\end{grammar}
\end{framed}\vspace{-1em}
\caption{Functions definitions}\vspace{2em}
\label{tab:t2_function_set}
\end{BNF}

\begin{BNF}
\setlength{\grammarindent}{6em}
\begin{framed}
\begin{grammar}
<section> ::= mov ax, timestamp\\
mov <reg>, 1664525\\
mul <reg>\\
add ax, 1013904223 <section>

<section> ::= mov <reg>, randint(0, 65,535)\\
mov <reg>, randint(0, 65,535)\\
xor <reg>, <reg>\\
shl <reg>, 7\\
shr <reg>, 5\\
xor <reg>, <reg> <section>
\end{grammar}
\end{framed}\vspace{-1em}
\setlength{\grammarindent}{2em}
\caption{Functions definitions for the random patterns.}\vspace{2em}
\label{table3_random_set}
\end{BNF}

\small

\bibliographystyle{apalike}
\bibliography{references,achiya}

\end{document}